%% file: root.tex
\documentclass[runningheads]{llncs}

\usepackage{eccv}

\usepackage{eccvabbrv}

\usepackage{graphicx}
\usepackage{booktabs}

\usepackage[accsupp]{axessibility}  

\usepackage{hyperref}

\usepackage{orcidlink}

\usepackage{tikz}
\usepackage{comment}
\usepackage{amsmath,amssymb} 
\usepackage{color}
\usepackage{xspace}

\usepackage{tabularray} 
\usepackage{amssymb}
\usepackage{multirow}
\usepackage{colortbl}

\newcommand*\rgbb{\includegraphics[width=10pt]{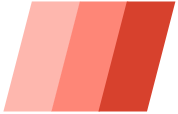}}
\newcommand*\nrgbb{\includegraphics[width=10pt]{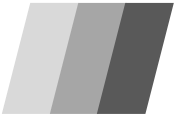}}
\newcommand*\avgg{\includegraphics[width=10pt]{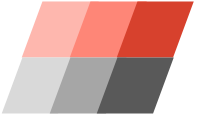}}
\newcommand{\rgb}{\raisebox{-0.1em}{\rgbb}}
\newcommand{\nrgb}{\raisebox{-0.1em}{\nrgbb}}
\newcommand{\avg}{\raisebox{-0.1em}{\avgg}}

\newcommand{\sysname}{\texttt{EdgeVL}\xspace}

\usepackage[framemethod=TikZ]{mdframed} 
\usepackage{dashrule} 
\usepackage{xcolor}
\definecolor{chatbox_color}{HTML}{FFFAED} 

\mdfdefinestyle{custombox}{%
    linecolor=black, 
    outerlinewidth=0.3pt, 
    roundcorner=2pt, 
    innertopmargin=\baselineskip, 
    innerbottommargin=\baselineskip, 
    innerrightmargin=50pt, 
    innerleftmargin=4pt, 
    backgroundcolor=chatbox_color, 
    frametitleaboveskip=\dimexpr-\ht\strutbox\relax, 
    font=\ttfamily\small, 
}

\begin{document}

\title{Self-Adapting Large Visual-Language Models to Edge Devices across Visual Modalities} 

\titlerunning{Self-Adapting Large VL Models to Edge Devices across Visual Modalities}

\author{
    Kaiwen Cai\inst{1}\orcidlink{0000-0001-7988-837X} \and
    Zhekai Duan\inst{1} \orcidlink{0000-0002-0283-8419} \and
    Gaowen Liu\inst{2} \and
    Charles Fleming\inst{2} \and
    \mbox{Chris Xiaoxuan Lu}\inst{3}\thanks{Corresponding author. Email: xiaoxuan.lu@ucl.ac.uk}\orcidlink{0000-0002-3733-4480}
}

\authorrunning{K.~Cai et al.}

\institute{
University of Edinburgh, UK \and
Cisco Research, USA \and
University College London, UK 
}

\maketitle
\input{sections/abstract.tex}
\input{sections/introduction_v2.tex}

\input{sections/related.tex}
\input{sections/method.tex}

\input{sections/experiments.tex}
\input{sections/conclusion}

%
%

\appendix
\section*{Appendix}

\input{sections/quantization}

\input{sections/superset}
\input{sections/results}

\bibliographystyle{splncs04}
\bibliography{egbib}
\end{document}

%% file: sections/abstract.tex
\begin{abstract}
Recent advancements in Vision-Language (VL) models have sparked interest in their deployment on edge devices, yet challenges in handling diverse visual modalities, manual annotation, and computational constraints remain. We introduce \sysname, a novel framework that bridges this gap by seamlessly integrating dual-modality knowledge distillation and quantization-aware contrastive learning. This approach enables the adaptation of large VL models, like CLIP, for efficient use with both RGB and non-RGB images on resource-limited devices without the need for manual annotations. \sysname not only transfers visual language alignment capabilities to compact models but also maintains feature quality post-quantization, significantly enhancing open-vocabulary classification performance across various visual modalities. Our work represents the first systematic effort to adapt large VL models for edge deployment, showcasing up to $15.4\%$ accuracy improvements on multiple datasets and up to 93-fold reduction in model size. Code available at \url{https://github.com/ramdrop/edgevl}.
\end{abstract}

%% file: sections/introduction_v2.tex
\section{Introduction}
In recent years, there has been a surge of interest in the development of Vision-Language (VL) models capable of conducting integrated reasoning across visual and textual data. Prominent large-scale VL models, such as CLIP \cite{radford2021clip}, typically employ distinct visual and text encoders. These encoders embed images and text into a common feature space, enabling direct comparison across two modalities. By evaluating the degree of similarity between the image embeddings and various potential text embeddings, these models facilitate zero-shot and open-vocabulary visual recognition, including image classification \cite{he2022openvocabulary}, semantic segmentation \cite{ghiasi2022scaling}, and object detection \cite{minderer2022simple}.

However, three significant challenges hinder the deployment of VL models on edge devices: (i) generalization to diverse visual modalities, (ii) label scarcity in the wild settings, and (iii) on-device resource limitation. \emph{Firstly}, an edge device often comes equipped with multiple sensors beyond standard RGB cameras, such as depth sensors and infrared cameras. These are indispensable in edge devices like field robots or smart doorbells for visual comprehension under challenging lighting conditions, like darkness, smoke, or fog. Despite this, the visual encoders in most large VL models are predominantly tailored to RGB images, leaving the adaptability of these models to alternative inputs such as depth or infrared images largely unexplored. \emph{Secondly}, while edge devices can generate a vast amount of images, these images are often unlabeled, presenting a significant obstacle in the wild where human-specified annotations are unavailable. This absence of labels prevents the straightforward application of model fine-tuning on annotated datasets. \emph{Thirdly}, even if the transfer of knowledge across different visual modalities becomes achievable, the substantial computational requirements of the visual encoders (\eg, ViT \cite{dosovitskiy2020an} used by CLIP \cite{radford2021clip}) render them impractical for edge devices, which are typically constrained by limited memory and TOPS (Tera Operations Per Second) performance.

\begin{figure}[!t]
    \centering
    \includegraphics[width=\linewidth]{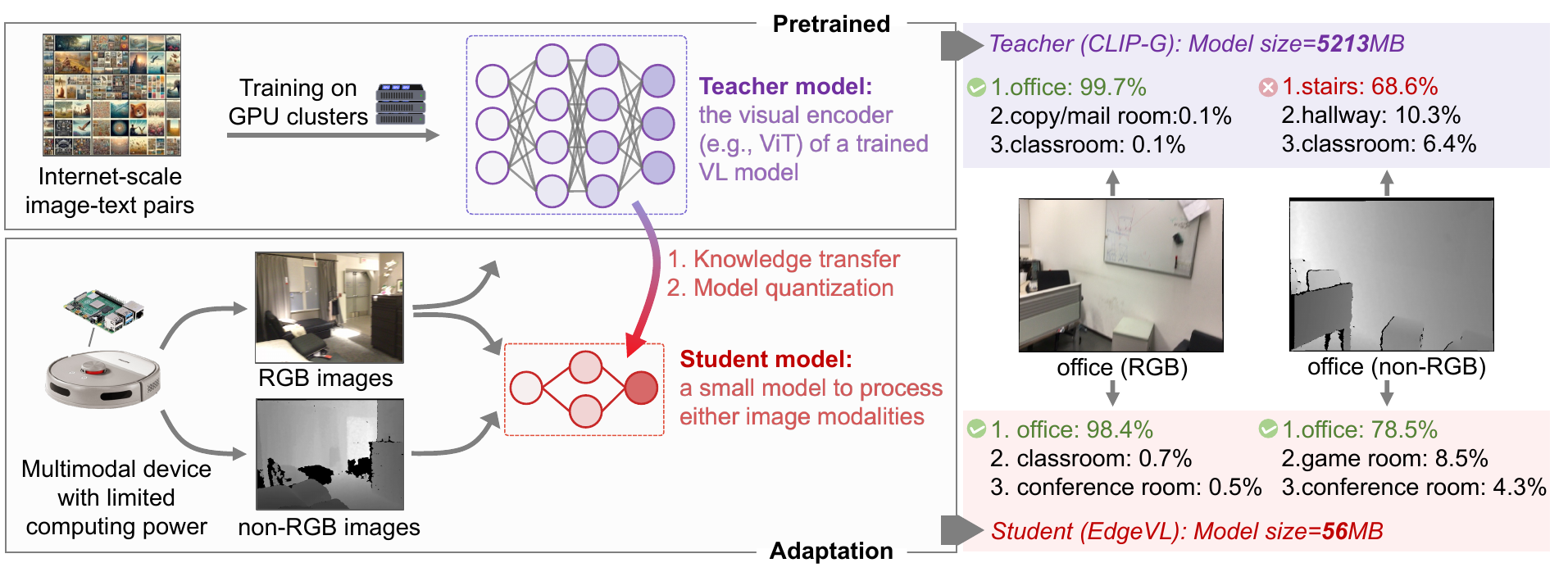}
    \caption{The adaptation problem of large visual language model to edge devices across visual modalities. We use a resource-constrained cleaning robot as the edge device for illustration. The robot has a \emph{co-located} RGB and depth cameras, generating many paired images without scene labels. Using RGB-depth pairs as the inputs and the pre-trained image encoder in CLIP as the teacher, \sysname is designed to transfer the knowledge to a small student encoder without labels or human intervention. After this learning process, the student encoder can agnostically process \emph{either} image modalities for open-vocabulary scene classification on the device.}
    \label{fig:openfig}
\end{figure}

To overcome these challenges, a novel framework is needed that can adapt the VL embedding prowess of large models to non-RGB images without relying on human annotations, while also minimizing its computational footprint to suit the capabilities of edge devices.
Existing literature tends to address these domains in isolation, focusing either on cross-modal knowledge transfer \cite{hong2022crossmodality, thoker2019crossmodal, hafner2022crossmodal} or on model compression (\eg, quantization \cite{jacob2018quantization}, pruning \cite{li2022revisiting} and distillation \cite{touvron2021training}). However, the interplay and potential synergy between these two areas remain largely unexplored, not to mention the impact of label scarcity. As evident in our empirical results (see \cref{tab:open-vocabulary-classification-quantized}), brute-forcing integrating the two modules leads to an obvious performance drop for large VL models. 

In this work, we propose \sysname, a streamlined two-stage adaptation framework that seamlessly integrates knowledge transfer with model compression. Initially, \sysname utilizes a dual-modality knowledge distillation process, leveraging a pre-trained visual encoder as the teacher model. This process distils knowledge to a more compact student model. The student model is designed to handle both RGB and non-RGB images, ensuring the alignment of visual features with textual representations similar to those found in conventional large VL models. This initial stage significantly enhances model efficiency through architectural optimization. Subsequently, to further boost efficiency and the efficacy of extracted features for edge deployment, the framework applies quantization-aware training (QAT) augmented with a novel contrastive learning loss. This sophisticated approach culminates in a low-bitrate visual encoder model, optimized for edge devices, which demonstrates superior performance in open-vocabulary classification tasks for both RGB and non-RGB images. \cref{fig:openfig} depicts the conceptual framework of \sysname. We summarise our main contributions as follows:
\begin{itemize}
\item \sysname is the first framework to systematically address the adaptation of large VL models for edge devices, facilitating their use with diverse visual modalities without relying on manual annotations.
\item We introduce a method to transfer visual language alignment from pre-trained VL models to compact visual models for both RGB and non-RGB images, eliminating annotation needs.
\item We incorporate quantization-aware training enhanced by a contrastive learning loss. This approach not only maintains the quality of feature representation post-quantization but also significantly improves the model's discriminative ability across diverse visual modalities.
\item We highlight \sysname's gain in accuracy across multiple datasets and detail its efficiency improvements on diverse GPU tiers. 
\end{itemize}

%% file: sections/related.tex
\section{Related Work}

\subsection{Open Vocabulary Classification}
Recent advances in VL models \cite{radford2021clip, jia2021scaling} have made it possible to obtain consistent image and text representations. In open vocabulary object detection tasks \cite{zareian2021openvocabulary, gu2022openvocabulary, du2022learning}, the classification head is replaced with a feature projection head, whose output is compared with the text embeddings of all candidate classes to obtain the final prediction. In open vocabulary semantic segmentation, FreeSeg \cite{qin2023freeseg} includes a task-specific prompt in the text queries and trains an encoder to be aligned with the CLIP text embeddings. OpenScene \cite{peng2023openscene} trained a 3D model to learn point embeddings that mimic the pixel embeddings from the CLIP image encoder. Although these open vocabulary classification methods \cite{kuo2023openvocabulary, ding2023pla, huynh2022openvocabulary, he2023clips} have achieved promising results, they are focusing on the RGB image modality. In this work, we aim to address the open vocabulary scene classification problem on various modalities beyond RGB images.

\subsection{Cross-modality Knowledge Distillation}
Knowledge distillation facilitates the transfer of expertise between models through various strategies such as response distillation \cite{hinton2015distilling}, feature distillation \cite{romero2015fitnets}, and relation distillation \cite{dai2021general}. In the realm of cross-modality knowledge distillation, methods like those presented by \cite{hoffman2016crossmodal} and \cite{hafner2022crossmodal} transfer knowledge by sharing weights between models of different modalities. \cite{thoker2019crossmodal} employs a strategy where multiple student models are encouraged to align their predictions with a teacher model, enhancing cross-modal action recognition. CMKD \cite{hong2022crossmodality} introduces a combination of feature and response distillation to propagate knowledge from the LiDAR modality to the RGB modality. Recently, CLIP has also been explored for transferring knowledge from 2D images to 3D scenes \cite{zhang2022pointclip, chen2023clipscene, xue2023ulip}. Unlike these methods tailored for labeled datasets, our work uniquely addresses the challenge of knowledge distillation using unlabeled pairs of RGB and non-RGB images.

\subsection{Model Quantization}

Model quantization techniques fall into two main categories: Post Training Quantization (PTQ) and Quantization Aware Training (QAT). PTQ methods, such as those outlined by \cite{nagel2019datafree, cai2020zeroq,dettmers2022gptint,xiao2023smoothquant,liu2023pdquant}, involve quantizing the weights and activations of a model after training, without retraining the model. For instance, LLM.int8 \cite{dettmers2022gptint} focuses on identifying weight outliers and applying distinct quantization methods for inliers and outliers, while SmoothQuant \cite{xiao2023smoothquant} adjusts the scale of weights and activations to maintain quantization precision for both outliers and inliers. When training data or validation data is not accessible, ZeroQ \cite{cai2020zeroq} synthesizes input data using Batch Normalization layer statistics to evaluate layer sensitivity. On the contrary, QAT, as explored by \cite{choi2018pact,jacob2018quantization,esser2020lsq,lee2021ewgs,krishnamoorthi2018quantizing}, is employed when PTQ does not satisfy accuracy requirements. This approach includes inserting fake quantization nodes during training, allowing the model to adapt to quantization errors and learn a more robust representation. Notably, LSQ \cite{esser2020lsq} improves quantization performance through a learnable scale factor, and EWGS \cite{lee2021ewgs} enhances gradient estimation with a weighted gradient scaling method.
In this work, we investigate 
quantization strategy for transferring cross-modal knowledge from large VL models.

%% file: sections/method.tex
\section{Methodology}
\subsection{Preliminary on Open Vocabulary Classification}

Large-scale VL models like CLIP~\cite{radford2021clip}, which include image and text encoders, are trained on over 400 million image-text pairs to map them into a shared feature space, optimizing true pair closeness and false pair distance through contrastive training. This approach enables CLIP to perform zero-shot and open-vocabulary classifications on unseen class labels during inference by assessing the similarity between image and text embeddings. However, despite its success with RGB images, CLIP's visual encoder underperforms in zero-shot classification tasks with non-RGB images. Performance data (see \cref{tab:open-vocabulary-classification-overall} later) indicates a significant accuracy disparity between RGB and non-RGB images (\eg, Depth and Infrared (IR)), with an example from the ScanNet dataset showing an approximately 8-fold decrease in accuracy for depth images compared to RGB.


\subsection{Problem Definition} 
\begin{figure}[!t]
    \centering
    \includegraphics[width=0.9\linewidth]{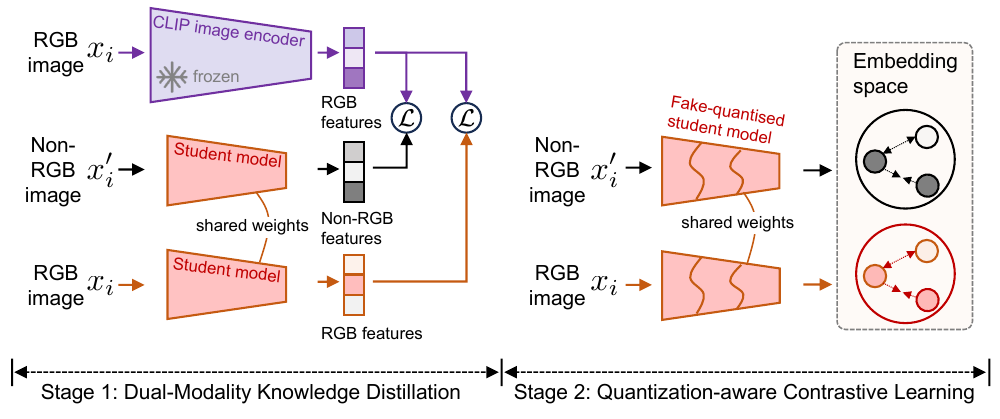}
    \caption{Overall architecture of our proposed method. In stage-1, we distill the knowledge from the pre-trained visual encoder to the student model. In stage-2, we first fake-quantize the pretrained student model, then use contrastive learning to refine the student model.}
    \label{fig_overall_architecture}
\end{figure}

Motivated by the preliminary, we consider adapting an open-vocabulary classifier to edge devices with different image sensors. Let the training set for adaptation be denoted by $\mathcal{D}_{train}=\{(x_i, x_i^\prime)\}_{i=1}^{N}$, where $x_i$ represents the $i^{th}$ RGB image, $x_i^\prime$ is its non-RGB image counterpart. Co-located cameras continuously collect the above pair of images on an edge device (\eg, a mobile robot), and we use $N$ pairs of such images for model adaptation. Note that, there are no labels for these images. Assume an RGB image encoder $\varPhi_{{img}}$ from a pre-trained large VL model is available. The adaptation goal for \sysname is to roll out a modality-agnostic and efficient image encoder $\varPhi_{{img}}^{{edge}}$ so that the following visual features can be approximately the same:
\begin{equation}
    \begin{aligned}
        \varPhi_{{img}}(x_i)  \approx \varPhi_{{img}}^{{edge}}(x_i) \approx \varPhi_{{img}}^{{edge}}(x_i^\prime)
    \end{aligned}
\end{equation}
The entire training of \sysname does not need human annotation or manual labels.

When it comes to the inference stage, assume there is a test set denoted as $\mathcal{D}_{test}=\{(x_i, x_i^\prime), y_i \in \mathcal{C}\}_{i=1}^{N}$, where $y_i$ designates the class label for each image pair, and $\mathcal{C}$ encompasses all possible open-vocabulary classes. 
By using the text encoder of the same pre-trained large VL model and the developed image encoder $\varPhi_{{img}}^{{edge}}$, the open-vocabulary class prediction can be translated into the maximum feature similarity inference problem:
\begin{equation}
\hat{y}_i = \underset{Y \in \mathcal{C}}{\arg\max} \; \varPhi_{{img}}^{{edge}}(x_i)^\top \varPhi_{{text}}(Y), \quad \hat{y}_i' = \underset{Y \in \mathcal{C}}{\arg\max} \; \varPhi_{{img}}^{{edge}}(x_i')^\top \varPhi_{{text}}(Y)
\end{equation}
Ideally, if $\varPhi_{{img}}^{{edge}}$ is a well-adapted image encoder, the predicted classes $\hat{y}_i$ and $\hat{y}_i'$ align closely with the true classes $y_i$ and the inference efficiency is boosted
\footnote{Class prediction varies by case, but classes are typically chosen for specific downstream tasks. Take the example of a mobile robot performing perception tasks: the image is the only live input, while class labels and \(\Phi_{{text}}(Y)\) features are predetermined. Thus, image encoding is the main computation bottleneck in our inference.}.
As illustrated in \cref{fig_overall_architecture}, \sysname consists of a two-stage adaptation framework progressively solving the above problem: i. dual-modality knowledge distillation module ($\varPhi_{{img}} \rightarrow \varPhi_{{img}}^{{stu}}$), and ii. quantization-aware contrastive learning module ($\varPhi_{{img}}^{{stu}} \rightarrow \varPhi_{{img}}^{{edge}}$).

\subsection{Stage-1: Dual-Modality Knowledge Distillation}
\label{sec_knowledge_distillation} 
The first stage of \sysname aims to distill the image features from the teacher image encoder in a pre-trained VL model to a student image encoder for dual modalities. 
For a trained large VL model like CLIP, while its image encoder has certain zero-shot transferability on unseen data, it still has failure cases where it provides noninformative or noisy features as supervision signals.
Removing the noisy samples and their impact on the distillation is beneficial, but the challenge arises when manual sample checking is costly in the wild.

\subsubsection{Automatic Dataset Curation}
We thus introduce an automated data selection mechanism as a precursor to feature distillation, guided by the innate comparison capabilities of VL models for images and texts. This approach leverages the VL models' ability to generate auxiliary information beneficial for sample selection without human intervention. Our approach initiates by creating a `superset of labels', $\mathcal{S}$, through a ChatGPT-4 \cite{achiam2023gpt} engine. 
Owing to the open-vocabulary feature of VL models, this label superset acts as an extensive repository of potential labels, applicable across various contexts. 
(See supplementary for the generated label superset)

We next encode this label superset into text features using a text encoder, $\varPhi_{{text}}$. Concurrently, an image encoder, $\varPhi_{{img}}$, processes unlabelled RGB images from edge devices to extract visual features. For an RGB image $x_i$, we assign a confidence score $c_i$ based on the highest image-text similarity with texts in $\mathcal{S}$:
\begin{equation}
    c_i = \max\{ s_k \mid s_k= \frac{e^{\varPhi_{{img}}(x_i)^\top\varPhi_{text}(y_k)}}{\sum_k^{\vert \mathcal{S}\vert} e^{\varPhi_{{img}}(x_i)^\top\varPhi_{text}(y_k)} }, k=1,2,\ldots,\vert \mathcal{S}\vert \},
\end{equation}
Our observations reveal that images with noisier and less informative features typically yield lower confidence scores.
Such images are deemed unsuitable for feature distillation and are excluded from further processing.
Concretely, we employ a predefined threshold, $\tau_{c}$, to curate the dataset. Only RGB images scoring above this threshold are retained in the training dataset, $\mathcal{D}_{train}=\{(x_i, x_i^\prime)\}_{i=1}^{N_{c}}$, which also includes their non-RGB counterparts collected simultaneously by the device. For notation simplicity, the training set $\mathcal{D}_{train}$ hereafter refers to the automatically curated dataset.

\subsubsection{Feature Distillation} 
With the curated dataset, we are poised to proceed with the feature distillation process. This crucial phase empowers a compact student encoder (\eg, Swin-T \cite{liu2021swin}) to effectively extract robust image embeddings from both RGB and non-RGB images, via referencing the large teacher encoder (\eg, ViT-T in CLIP \cite{radford2021clip}). Our approach diverges from conventional methods that tailor a compact student model solely for non-RGB or RGB image inputs. Instead, we advocate for a unified image encoder capable of seamlessly processing either image type through weight sharing. This innovative dual-modality student encoder not only simplifies the model architecture but also significantly reduces model storage requirements by at least half on edge devices.
Illustrated in Figure \ref{fig_overall_architecture}, for each pair of RGB and non-RGB images in $\mathcal{D}_{train}$, we align the features extracted from both RGB and non-RGB images by the student model with those extracted from RGB images by the teacher image encoder. This alignment is predicated on the understanding that both image types represent the same scene, thereby necessitating that the student model generates consistent image features that resonate with those from a pre-trained VL model (i.e., teacher) like CLIP. We designate the student encoder as $\varPhi_{img}^{stu}$ and focus on minimizing the discrepancy between the student model's features and the teacher's image features through our feature distillation loss function:
\begin{equation}
    \begin{aligned}
        \mathcal{L}_d = d(\varPhi_{img}({x}), \varPhi_{img}^{stu}({x}^\prime)) + d(\varPhi_{img}({x}), \varPhi_{img}^{stu}({x})).
    \end{aligned}
    \label{eq:stage1_loss}
\end{equation}
Here, $d$ signifies the distance function. We follow \cite{fang2023simple} and use the L1 distance function.
Through this loss function, we aim to closely align the student's feature representations with those of the teacher model, thereby ensuring the student model's proficiency across both RGB and non-RGB modalities.

\subsection{Stage-2: Quantization-aware Contrastive Learning}
\label{sec_quantization_aware_contrastive_learning}

Given the dual-modality student encoder $\varPhi_{img}^{stu}$, the next step is to further enhance its efficiency by transforming it to a quantized low-bit model $\varPhi_{img}^{edge}$ on par with the resources available on edge devices. The challenge, however, is how to preserve the feature expressiveness after quantization is applied.

\subsubsection{QAT Meets Contrastive Learning}

\begin{figure}[!tp]
    \centering
    \includegraphics[width=0.9\linewidth]{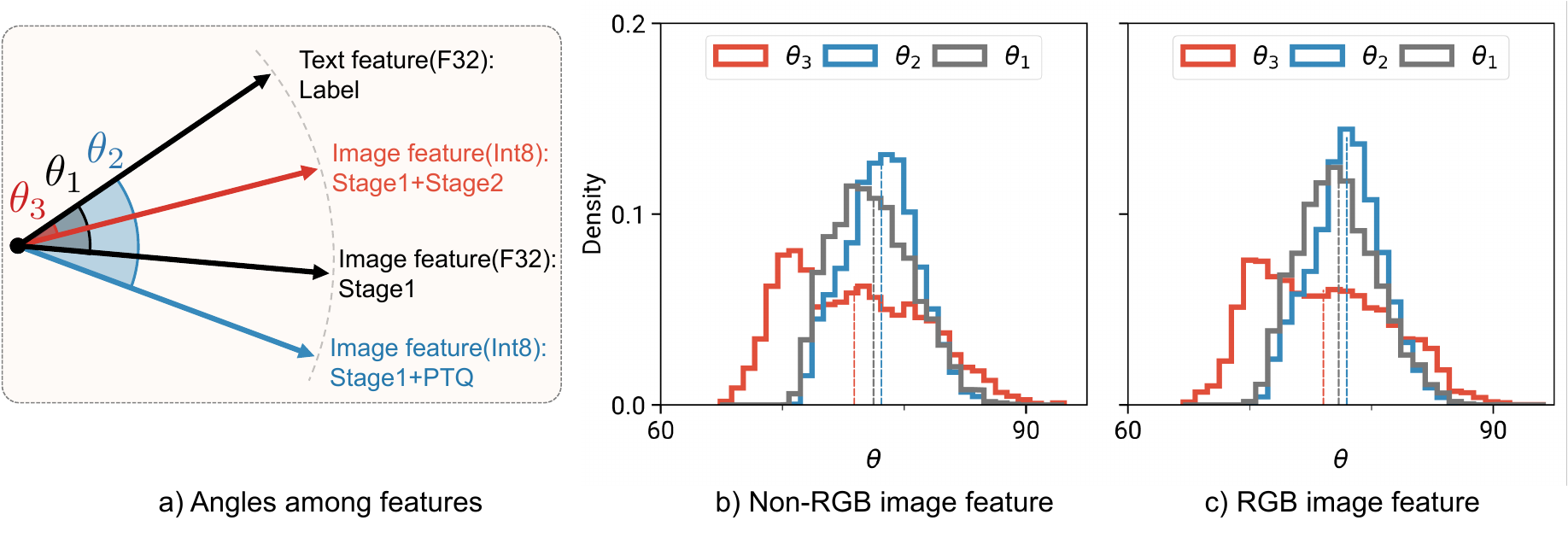}
    \caption{Angles between the features of images and their corresponding text labels on the ScanNet dataset: We calculate the angles based on the cosine similarities (a lower cosine similarity corresponds to a greater angle between features \cite{han2022data}). A rightward shift in the angle distribution in b) and c) suggests that \(\theta_2 > \theta_1\), indicating that image features diverge from the text labels following PTQ. Conversely, a leftward shift implies \(\theta_3 < \theta_1\), showing that image features align more closely with the text labels after Stage 2. Dashed lines denote mean values. Best viewed in color.}
    \label{fig_angles}
\end{figure}

To explore the impact of model quantization on the expressiveness of features, we initially apply PTQ~\cite{jacob2018quantization} on $\varPhi_{{img}}^{{stu}}$ and examine the discriminative nature of features post-quantization. As demonstrated in \cref{fig_angles}, the discriminability of features notably declines after quantization, leading to a misalignment between text and image representations. This reduction in feature clarity compared to the full-precision model underscores the necessity of employing QAT for the final image encoder optimization. QAT enhances the model by incorporating fake quantization during training, which simulates quantization effects through quantization-aware matrix multiplication, followed by finetuning the pre-trained model to adapt to these effects (see supplementary for details of QAT).
A pivotal aspect of implementing QAT is selecting an appropriate loss function that maintains or even improves the discriminative capability of features within the quantization framework. Traditional knowledge distillation loss, as formulated in \cref{eq:stage1_loss}, aims to align the student model's features with those of a pre-trained teacher model. However, it might not exploit the full potential of quantized models in achieving robust and discriminative features. In this context, we propose the integration of contrastive learning loss, which is designed to cultivate representations resilient to non-discriminative features and enhances the separation in the feature space between similar and dissimilar instances. 
This approach is predicated on the robustness of contrastive learning, which should support the acquisition of invariant representations less affected by the distortions due to quantization. \cref{fig_angles} illustrates how employing contrastive learning loss in conjunction with QAT not only mitigates the discriminative power reduction but potentially increases the discriminability of the student encoder after quantization.

\subsubsection{Triplet Sampling}
Selecting positive and negative samples effectively is pivotal for conservative learning. 
We adopt
the semi-hard sample strategy \cite{chen2020simple}, recognized for its capacity to improve feature robustness, in developing a conservative learning loss. Specifically, for each pair of samples, $(x_i, x_i^\prime)$, within the training dataset $\mathcal{D}_{train}$, we generate pseudo labels from the superset $\mathcal{S}$ by utilizing a pre-trained VL model as follows:
\begin{equation}
    \begin{aligned}
        \hat{y}_i &= \underset{Y \in \mathcal{S}}{\arg\max} \; \varPhi_{img}(x_i)^\top \varPhi_{{text}}(Y).
    \end{aligned}
\end{equation}
Then, for each training instance $x_i$, we identify its corresponding potential positive samples $\{p_{i, k}\}$ and potential negative samples $\{n_{i,j}\}$. Here, potential positives are those samples sharing the same pseudo label as $x_i$, and potential negatives are those with differing pseudo labels.
We select the optimal matching positive sample $p_{i, k^*}$ by:
\begin{equation}
    \label{eq:optimal_matching}
    \begin{aligned}
        k^* = \underset{k}{\arg\min} \; d(\varPhi_{{img}}^{{edge}}(x_i), \varPhi_{{img}}^{{edge}}(p_{i, k})),
    \end{aligned}
\end{equation}
and randomly choose negatives. We subsequently retain only those negative samples such that the semi-hard condition is met:

\begin{equation}
    \left\{
    \begin{aligned}
        & d(\varPhi_{{img}}^{{edge}}(x_i), \varPhi_{{img}}^{{edge}}(n_{i,j})) > d(\varPhi_{{img}}^{{edge}}(x_i), \varPhi_{{img}}^{{edge}}(p_{i, k^*})),  \\
        & d(\varPhi_{{img}}^{{edge}}(x_i), \varPhi_{{img}}^{{edge}}(n_{i,j})) < d(\varPhi_{{img}}^{{edge}}(x_i), \varPhi_{{img}}^{{edge}}(p_{i, k^*})) + m,
    \end{aligned}
    \right.
\end{equation}
where $m$ is a predefined constant margin. Denoting the size of the refined negative set as $J$, we define the loss function:
\begin{equation}
    \begin{aligned}
        \mathcal{L}_c = \frac{1}{J}\sum_{j=1}^{J} d(f(x_i), f(p_{i, k^*})) - d(f(x_i), f(n_{i,j})) +m. 
    \end{aligned}
\end{equation}


%% file: sections/experiments.tex
\section{Experimental Results}

\subsection{Implementation}
We use the CLIP model, ViT-g-14 (ViT-G), provided by OpenCLIP \cite{ilharco_gabriel_2021_5143773} as the teacher model. The student model is built upon ViT-S and its \textit{state-of-the-art} (SOTA) variants, DAT-T \cite{xia2022vision} and Swin-T \cite{liu2021swin}, where we replace the classification head with a feature projection head. For training with student models in stage 1, we use AdamW optimizer with a base learning rate of ${10^{-4}}$ and weight decay of 0.05. We use a cosine learning rate scheduler which decays the learning rate to $5\times{10^{-6}}$ over $120$ epochs. In stage 2, we reduce the base learning rate to ${10^{-6}}$. 
For the threshold $\tau_c$, we empirically set it at $0.25$ considering the training data utilization and data noise. 
For the CLIP text encoder, we use the text prompt ``a photo of a \{scene category\}." or ``a satellite image of a \{scene category\}." depending on the dataset as suggested by \cite{radford2021clip}. 
For quantized models, we report static quantization results. 
For the triplet loss $\mathcal{L} _c$, we use a margin $m=0.3$ and a negative set size $J=3$.

\subsection{Overall Results}
\label{subs:overall_results}
The overall results assess \sysname against several SOTA baseline methods on different datasets. We focus on two pivotal metrics: accuracy and efficiency. These metrics are crucial for demonstrating the effectiveness and practical applicability of our method on edge devices.

\subsubsection{Datasets}
The ScanNet and EuroSAT datasets are selected for evaluation. The \textit{ScanNet} dataset \cite{dai2017scannet} includes 1.89M training, 0.53M validation, and 0.21M test indoor RGB-D images. Adhering to \cite{fischedick2023efficient}, we applied a subsampling factor of 100 to the dataset to diminish the presence of similar images, resulting in a dataset comprising 18,900 training, 5,300 validation, and 2,100 test RGB-D images across 21 scene categories. Since the test split does not provide labels, we evaluate models on the validation split. The \textit{EuroSAT} dataset \cite{helber2018eurosat}, providing satellite images across 13 spectral bands and encompassing 10 classes with a total of 27,000 images, was randomly split into 13,500 training and 13,500 testing divisions. For the evaluation of cross-modality performance, we utilized the RGB and SWIR (Short Wave Infrared) bands, ensuring a comprehensive assessment of our method's effectiveness across varied imaging conditions.

\subsubsection{Baselines}
Given the absence of a direct baseline for our novel problem setting, we adapt several methods addressing similar challenges to fit our context. The first adapted baseline, \textit{CMKD} \cite{hong2022crossmodality}, was initially designed for transferring knowledge from LiDAR to RGB models. We modify it for distilling knowledge from the CLIP visual encoder to both RGB and non-RGB models. The \textit{Fida} framework \cite{thoker2019crossmodal}, which implements a dual student model approach within a mutual teacher-student learning paradigm, is tailored to our needs by focusing on minimizing the feature distance between student model pairs. Similarly, the \textit{CQD} method \cite{su2017adapting}, originally aimed at knowledge distillation from high-resolution to low-resolution models, is redirected towards reducing the feature distance between non-RGB models and both the pre-trained RGB model and the CLIP visual encoder. The \textit{SKD} strategy \cite{yang2022mixskd}, sharing conceptual similarities with our work through its mixup technique for generating hybrid-modality samples, is specifically adapted by us to integrate non-RGB and RGB images for training purposes. Lastly, the \textit{Frank} approach \cite{hafner2022crossmodal} and the \textit{Gupta} technique \cite{hoffman2016crossmodal} are considered for their relevance in cross-modal weight transfer and fine-tuning, and developing modality-specific models that converge through a unified embedding layer for efficient multimodal data processing, respectively. 

To showcase the efficacy of \sysname, we present the optimal outcomes for baseline models by employing their full-precision (F32) configurations and selecting the backbone that yields the highest accuracy for each. More detailed comparisons with the baselines per backbone can be found in the supplementary. 
Our comparison also includes two different versions of CLIP with ViT-B/G visual backbones respectively. We term them CLIP-B and CLIP-G hereafter.   
\subsubsection{Accuracy}

\begin{figure}[t]
  \centering
  \includegraphics[width=0.9\linewidth]{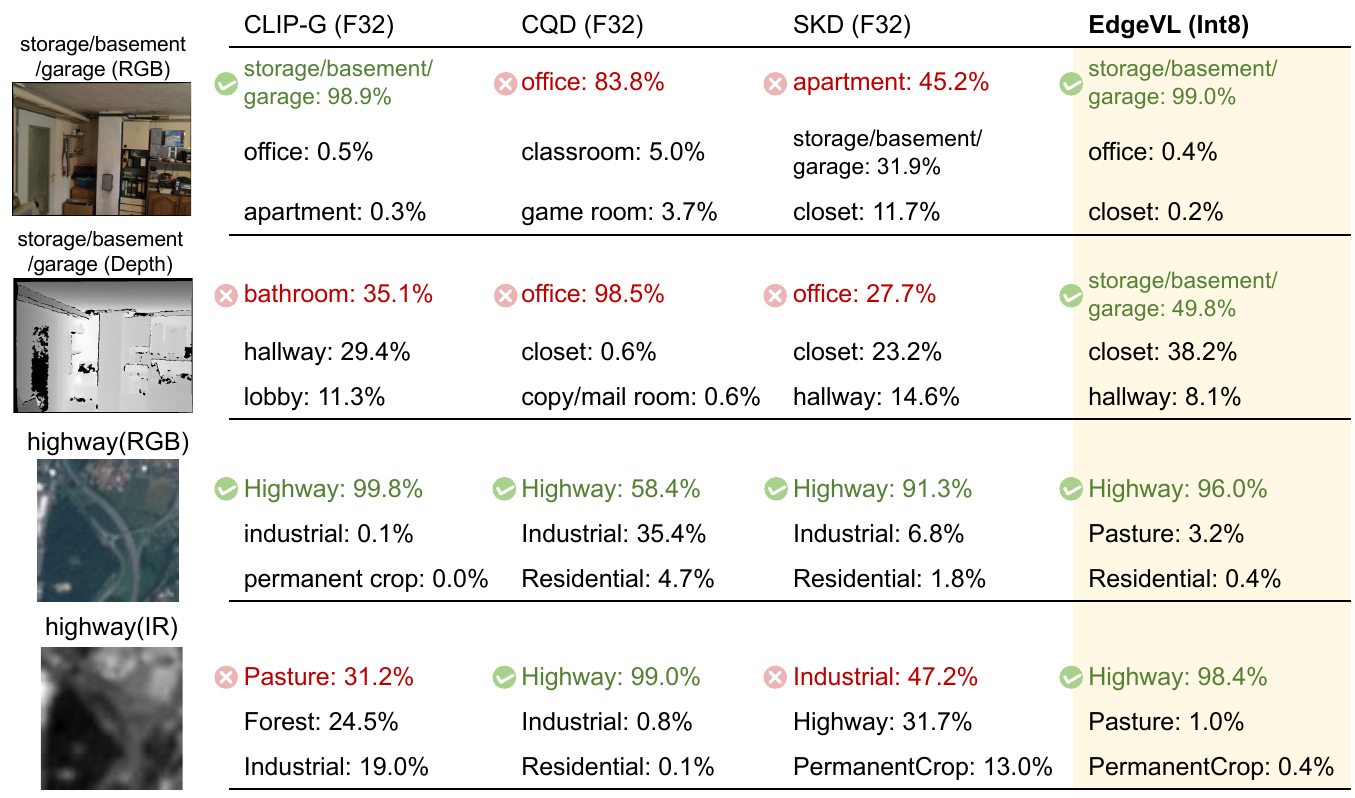}
  \caption{Visualization of the predictions of different models on ScanNet and EuroSAT. CLIP-G, CQD \cite{su2017adapting} and SKD \cite{yang2022mixskd} fall short for non-RGB images, while \sysname(Swin-T) demonstrates superior performance across both image modalities.}
  \label{fig:visual_examples}
\end{figure}

\cref{tab:open-vocabulary-classification-overall} displays the accuracy of \sysname in comparison to baseline models across the ScanNet and EuroSAT datasets, illustrating that \sysname secures the highest accuracy on both. Notably, the least performing variant of \sysname, ViT-S, significantly outperforms its closest rival, SKD\cite{yang2022mixskd}, by an impressive margin of $10.2\%$ on ScanNet (34.5\% vs. 44.7\%). This gap widens to $13.9\%$ against the top baseline, CQD\cite{su2017adapting}, on EuroSAT (49.4\% vs. 64.8\%). 
The consistent outperformance of \sysname across various backbone architectures for student encoders underscores its adaptability and broad applicability. Last but not least, compared with two pre-trained CLIP models, \sysname demonstrates superior performance across individual accuracy metrics for both RGB and non-RGB images. This highlights the importance of VL adaptation in the target domain.
\cref{fig:visual_examples} further shows the qualitative results.

\begin{table}[t]
  \centering
  \renewcommand\arraystretch{1.2}
  \setlength\tabcolsep{4pt}  
  \caption{Overall accuracy comparison. \nrgb\ and \rgb\ denote the top1 accuracy of non-RGB and RGB images, respectively. And \avg\ denotes the average of the two, all in percentage. The same applies to the following tables. Best viewed in color.}
  \label{tab:open-vocabulary-classification-overall}
  \scalebox{0.87}{
  \begin{tabular}{l|l|ccc|ccc} 
  \hline
  \noalign{\vskip 2pt} 
  \multirow{2}{*}{Methods}                            & \multirow{2}{*}{Bits} & \multicolumn{3}{c|}{ScanNet (\%) $\uparrow$ }                        & \multicolumn{3}{c}{EuroSAT (\%) $\uparrow$}                          \\ 
                                                      &                       & \nrgb      & \rgb       & \avg            & \nrgb      & \rgb       & \avg             \\ 
                                                      \noalign{\vskip 2pt} 
                                                      \hline
  \noalign{\vskip 2pt} 
  Pretrained CLIP-B \cite{radford2021clip}                   & F32                   & 4.5          & 36.2          & 20.4          & 16.8          & 40.4          & 28.6           \\
  
  Pretrained CLIP-G \cite{radford2021clip}                   & F32                   & 6.2          & 47.3          & 26.8          & 16.9          & 54.0          & 35.5           \\
  \noalign{\vskip 2pt} 
  \hline
  \noalign{\vskip 2pt} 
  Frank \cite{hafner2022crossmodal}                   & F32                   & 8.3          & 21.7          & 15.0          & 49.2          & 37.9          & 43.5           \\
  Gupta \cite{hoffman2016crossmodal}                  & F32                   & 16.0          & 17.5          & 19.8          & 54.2          & 42.4          & 48.3           \\
  CMKD \cite{hong2022crossmodality} (non-RGB)           & F32                   & 37.8          & 11.5          & 24.6          & 61.2          & 34.4          & 47.8           \\
  CMKD \cite{hong2022crossmodality} (RGB)             & F32                   & 4.0          & 42.5          & 23.2          & 20.1          & 62.4          & 41.2           \\
  Fida \cite{thoker2019crossmodal}                    & F32                   & 38.9          & 5.8          & 22.3          & 56.7          & 20.3          & 38.5           \\
  CQD \cite{su2017adapting}                           & F32                   & 40.1          & 6.7          & 23.4          & 62.4          & 36.4          & 49.4           \\
  SKD \cite{yang2022mixskd}                           & F32                   & 31.2          & 37.8          & 34.5          & 22.9          & 50.3          & 36.6           \\
  \rowcolor[rgb]{1, 0.9725, 0.894} \sysname (DAT-T)  & Int8                  & \textbf{47.9} & \textbf{52.0} & \textbf{49.9} & 61.0          & 65.7          & 63.3           \\
  \rowcolor[rgb]{1, 0.9725, 0.894} \sysname (Swin-T) & Int8                  & 46.0          & 48.7          & 47.4          & 61.3          & \textbf{67.1} & 64.2           \\
  \rowcolor[rgb]{1, 0.9725, 0.894} \sysname (ViT-S)  & Int8                  & 42.0          & 47.5          & 44.7          & \textbf{62.9} & 66.8          & \textbf{64.8}  \\
  \noalign{\vskip 2pt} 
  \hline
  \end{tabular}
  }
\end{table}

\begin{table}[t]
  \centering
  \renewcommand\arraystretch{1.2}
  \setlength\tabcolsep{2pt}   
  \caption{Overall efficiency comparison on different GPU platforms}
  \label{tab:efficiency_comparison}
  \scalebox{0.85}{
  \begin{tabular}{l|c|c|cc|c} 
    \hline
    \noalign{\vskip 2pt} 
    \multirow{2}{*}{Methods}                       & \multirow{2}{*}{Bits} & \multirow{2}{*}{Model Size $\downarrow$ } & \multicolumn{2}{c|}{Latency $\downarrow$ } & Throughput $\uparrow$ \\ 
    \cline{4-6}
                                                   &                       &                                  & AGX   & Nano                       & RTX4090                \\ 
                                                   \noalign{\vskip 2pt} 
                                                   \hline
                                                   \noalign{\vskip 2pt} 
    Pretrained CLIP-G                              & F32                   & 5213 MB                             & /     & /                          & /                      \\
    Pretrained CLIP-B                              & F32                   & 330 MB                              & 9.5 ms& 20.2 ms                      & 772 image/s                    \\
    \rowcolor[rgb]{1,0.973,0.894} \sysname(ViT-S)  & Int8                  & 86 MB                               & 4.6 ms ($\downarrow 52\%$)  & 9.9 ms ($\downarrow 51\%$)                       & 1492 image/s ($\uparrow 93\%$)                  \\
    \rowcolor[rgb]{1,0.973,0.894} \sysname(Swin-T) & Int8                  & 56 MB                               & 5.2 ms ($\downarrow 46\%$) & 11.4 ms  ($\downarrow 44\%$)                    & 1098 image/s ($\uparrow 42\%$)                  \\
    \noalign{\vskip 2pt} 
    \hline
    \end{tabular}
  }
    \end{table}

\subsubsection{Efficiency} 
We examine the computational efficiency of \sysname on the following Nvidia GPUs: i. Jetson AGX Orin (32GB), ii. Jetson Orin Nano (8GB) iii. RTX4090. To simulate actual deployment scenarios, we conduct all model inferences using the TensorRT engine, excluding DAT-T from profiling due to its current incompatibility with TensorRT. As CLIP with ViT-G backbone (CLIP-G) is not compatible with TensorRT, we report its model size only.




\subsection{Ablation Study} 

\subsubsection{Quantization Aware Contrastive Learning}
For comparison, we employ \textit{+PTQ} \cite{wu2020integer}, which converts model weights and activations into 8-bit integers. The process of \textit{+QAT} \cite{jacob2018quantization} involves the integration of fake quantization layers within the model, followed by finetuning to refine its performance. Similarly, \textit{+QViT} \cite{li2022qvit} enhances model efficiency through a combination of information rectification and distribution-guided distillation during the finetuning stage.

\cref{tab:open-vocabulary-classification-quantized} demonstrates the open-vocabulary classification accuracy of int8 quantized models using different methods for edge device deployment, focusing on static quantization results, though dynamic quantization aligns similarly and is detailed in the supplementary. PTQ significantly lowers accuracy, whereas QAT with stage-1 loss \cref{eq:stage1_loss} improves it but doesn't reach the \sysname benchmark, sometimes lagging by over $9.7\%$ (40.3\% vs. 50.0\%). In comparison, stage-2 in \sysname introduces contrastive learning in the quantization-aware training process, which improves the student model's discriminability and accuracy despite quantization.
The result highlights \sysname's effectiveness and robustness in preserving model performance after quantization, outperforming other strategies including QViT and PTQ by significant margins. 

%

\begin{table}[!t]
  \centering
  \renewcommand\arraystretch{1.2}
  \setlength\tabcolsep{4pt}  
  \caption{QAT meets Contrastive Learning. + denotes in combination with stage-1.}
  \label{tab:open-vocabulary-classification-quantized}
  \scalebox{0.87}{
    \begin{tabular}{l|l|ccl|ccl|ccl} 
      \hline
      \noalign{\vskip 2pt} 
      \multirow{2}{*}{Methods}                        & \multirow{2}{*}{Bits} & \multicolumn{3}{c|}{DAT-T (\%) }                                & \multicolumn{3}{c|}{Swin-T (\%)}                               & \multicolumn{3}{c}{ViT-S (\%)}                                 \\
                                                      &                       & \nrgb         & \rgb          & \multicolumn{1}{c|}{\avg} & \nrgb         & \rgb          & \multicolumn{1}{c|}{\avg} & \nrgb         & \rgb          & \multicolumn{1}{c}{\avg}  \\ 
                                                      \noalign{\vskip 2pt} 
                                                      \hline
      \noalign{\vskip 2pt} 
      +Stage-1                                         & F32                   & 38.6          & 40.6          & 39.6                      & 39.9          & 41.2          & 40.5                      & 37.8          & 40.7          & 39.3                      \\ 
      \noalign{\vskip 2pt} 
      \hline
      \noalign{\vskip 2pt} 
      +PTQ~\cite{jacob2018quantization}               & Int8                  & 33.0          & 36.5          & 34.8                      & 29.0          & 31.7          & 30.3                      & 24.7          & 25.9          & 25.3                      \\
      +QAT \cite{jacob2018quantization}               & Int8                  & 39.4          & 41.2          & 40.3                      & 38.9          & 39.7          & 39.3                      & 37.7          & 41.1          & 39.4                      \\
      +QViT~\cite{li2022qvit}                         & Int8                  & 35.0          & 38.0          & 36.5                      & 36.5          & 38.5          & 37.5                      & 31.4          & 35.3          & 33.3                      \\
       \textbf{+Stage-2} & Int8                  & \textbf{47.9} & \textbf{52.0} & \textbf{50.0}             & \textbf{46.0} & \textbf{48.7} & \textbf{47.4}             & \textbf{42.0} & \textbf{47.5} & \textbf{44.7}             \\
      \noalign{\vskip 2pt} 
      \hline
      \end{tabular}
  }
  \end{table}
  
\subsubsection{Dual-modality Knowledge Distillation} 

We evaluate the outcomes of our stage-1 training against CMKD(non-RGB)\cite{hong2022crossmodality} and CMKD(RGB) \cite{hong2022crossmodality}, which resemble our stage-1 but rely on single-modality training. \cref{tab:effect_dual_modality} illustrates the accuracy of these methods using the DAT-T model.
Our findings show that dual-modality training draws on the strengths of RGB-modality and non-RGB-modality training, yielding high accuracy for both 
modalities, whether non-RGB are depth images from ScanNet or infrared images from EuroSAT. 
Our dual-modality training method achieves an increase of $15.0\%$ on ScanNet and $13.1\%$ on EuroSAT in average accuracy. Notably, the accuracy boosts on non-RGB modality (\eg, 38.6\% vs. 37.8\% and 61.5\% vs. 61.2\%) lead us to hypothesize that RGB images function as a type of data augmentation for non-RGB images in dual-modality training.


\subsubsection{Cutoff Ratio of $\tau_c$}
\cref{tab_cutoff_ratio} presents the accuracy of \sysname(Swin-T) on EuroSAT, where $\tau _c=0.25$ brings the highest accuracy compared to the other cutoff ratios. A very small $\tau _c$ fails to fully train the 
model, whereas a very large $\tau _c$ introduces 
noisy samples, thus 
diminishing 
the effectiveness of knowledge transfer.

\begin{table}[!t]
  \centering
  \renewcommand\arraystretch{1.2}
  \setlength\tabcolsep{4pt}  
  \caption{Effect of Dual-modality Knowledge Distillation}
  \label{tab:effect_dual_modality}
  \scalebox{0.8}{
  \begin{tabular}{l|l|ccc|ccc} 
  \hline
  \noalign{\vskip 2pt} 
  \multirow{2}{*}{Methods}                                & \multirow{2}{*}{Bits} & \multicolumn{3}{c|}{ScanNet (\%)}      & \multicolumn{3}{c}{EuroSAT (\%)}        \\ 
                                                          &                       & \nrgb & \rgb & \avg   & \nrgb & \rgb  & \avg    \\ 
                                                          \noalign{\vskip 2pt} 
                                                          \hline
  \noalign{\vskip 2pt} 
  CMKD \cite{hong2022crossmodality} (non-RGB)           & F32                   & 37.8          & 11.5          & 24.6          & 61.2          & 34.4          & 47.8           \\
  CMKD \cite{hong2022crossmodality} (RGB)             & F32                   & 4.0          & \textbf{42.5}          & 23.2          & 20.1          & \textbf{62.4}          & 41.2           \\  
  \rowcolor[rgb]{1, 0.9725, 0.894} Stage-1 (Dual-modality) & F32                   & \textbf{38.6}      & 40.6     & \textbf{39.6} & \textbf{61.5}      & 60.3     & \textbf{60.9}  \\
  \noalign{\vskip 2pt} 
  \hline
  \end{tabular}
  }
\end{table}

\begin{table}[tp]
  \begin{minipage}[t]{0.35\linewidth}
    \renewcommand\arraystretch{1.2}
    \setlength\tabcolsep{4pt}
    \centering
    \caption{Effect of varying $\tau _c$ 
    }
    \label{tab_cutoff_ratio}
    \resizebox{0.735\columnwidth}{!}{
    \begin{tabular}{c|ccc}     
      \hline
      \noalign{\vskip 2pt} 
      $\tau _c$  & \nrgb  & \rgb  & \avg   \\ 
      \noalign{\vskip 2pt} 
      \cline{2-4}         
      \hline
      \noalign{\vskip 2pt} 
      0.10 &  49.8     & 52.9      &  51.4 \\
      0.25 &  \textbf{61.3}           &    \textbf{67.1}        &    \textbf{64.2}    \\
      0.50 &  61.2           &    62.8        &    62.0    \\
      \noalign{\vskip 2pt} 
      \hline
      \end{tabular}
    }  
    \end{minipage}
    \hspace{0cm}
  \begin{minipage}[t]{0.3\linewidth}
    \renewcommand\arraystretch{1.2}
    \setlength\tabcolsep{4pt}
    \centering
    \caption{Effect of different sampling strategies 
    }
    \label{tab_sampling_strategy}
    \resizebox{\columnwidth}{!}{
      \begin{tabular}{l|ccc} 
        \hline
        \noalign{\vskip 2pt} 
        Sampling & \nrgb  & \rgb  & \avg   \\ 
        \noalign{\vskip 2pt} 
        \hline
        \noalign{\vskip 2pt} 
        Semi-hard & \textbf{61.0}       & \textbf{65.7}      & \textbf{63.3}  \\
        Hard & 60.8       & 64.7      & 62.7  \\
        \noalign{\vskip 2pt} 
        \hline
        \end{tabular}
    }  
    \end{minipage}  
    \hspace{0cm}
  \begin{minipage}[t]{0.3\linewidth}
  \renewcommand\arraystretch{1.2}
  \setlength\tabcolsep{4pt}
  \centering
  \caption{Effect of different training strategies
  }
  \label{tab_two_stage_training}
  \resizebox{\columnwidth}{!}{
    \begin{tabular}{l|ccc} 
      \hline
      \noalign{\vskip 2pt} 
      Training & \nrgb  & \rgb  & \avg   \\ 
      \noalign{\vskip 2pt} 
      \hline
      \noalign{\vskip 2pt} 
      Two-stage &   \textbf{47.9}     & \textbf{52.0}      &  \textbf{49.9} \\
      One-stage &  27.1           &    33.0        &    30.0    \\
      \noalign{\vskip 2pt} 
      \hline
      \end{tabular}
  }  
  \end{minipage}
  \vspace{-10pt}
\end{table}

\subsubsection{Triplet Sampling Strategy}
For positives and negatives mined according to \cref{eq:optimal_matching}, we can choose to build triplets that include: 1) hard triplets, where the anchor-positive distance is greater than the anchor-negative distance or 2) semi-hard triplets, where the anchor-negative distance is greater than the anchor-positive distance but less than the anchor-positive distance plus a margin $m$. We are assessing how different sampling strategies impact the performance of \sysname. \cref{tab_sampling_strategy} presents the accuracy of \sysname(DAT-T) with different sampling strategies on EuroSAT. It demonstrates that semi-hard strategy outperforms hard strategy. This is consistent with the findings in \cite{schrofffacenet}.

\subsubsection{Two-stage Training}
\sysname\ follows a two-stage training process. We initially explored a one-stage alternative, replacing QAT with PTQ during model quantization to prevent model collapse during training. \cref{tab_two_stage_training} shows the accuracy of \sysname(DAT-T) with different training strategies on  ScanNet. It can be observed that One-stage training generates much inferior accuracy than Two-stage, \ie, \sysname. We hypothesize that this is because contrastive learning benefits from a well-organized feature space as a starting point. Additionally, QAT training necessitates a small learning rate to prevent model collapse, while in our case knowledge distillation training is effective with relatively large learning rates.

\subsection{Cross-Dataset Performance}
\label{sub:cross_dataset_performance}

\begin{table}[!t]
  \centering
  \renewcommand\arraystretch{1.2}
  \setlength\tabcolsep{4pt}  
  \caption{Accuracy on unseen datasets with the image encoder adapted on ScanNet}
  \vspace{-5pt}
  \label{tab:cross-dataset}
  \scalebox{0.85}{
  \begin{tabular}{l|l|ccl|ccl} 
  \hline
  \noalign{\vskip 2pt} 
  \multirow{2}{*}{Methods}                           & \multirow{2}{*}{Bits} & \multicolumn{3}{c|}{NYU2 (\%)}                           & \multicolumn{3}{c}{SUNRGBD (\%)}                         \\ 
                                                     &                       & \nrgb  & \rgb  & \multicolumn{1}{c|}{\avg} & \nrgb  & \rgb  & \multicolumn{1}{c}{\avg}  \\ 
                                                     \noalign{\vskip 2pt} 
                                                     \hline
  \noalign{\vskip 2pt} 
  Pre-trained CLIP-G                                   & F32                   & 25.7       & \textbf{69.7}      & 47.7                    & 18.0       & \textbf{54.3}      & \textbf{36.2}                    \\
  Pre-trained CLIP-B                                   & F32                   & 22.6       & 62.2      & 42.4                    & 15.2       & 47.2      & 31.2                    \\  
  \rowcolor[rgb]{1, 0.9725, 0.894} \sysname: DAT-T  & Int8                  & \textbf{51.1}       & 54.3      & \textbf{52.7}                    & 28.6       & 31.8      & 30.2                    \\
  \rowcolor[rgb]{1, 0.9725, 0.894} \sysname: Swin-T & Int8                  & 43.4       & 43.3      & 43.4                    & \textbf{30.0}       & 31.4      & 30.7                    \\
  \rowcolor[rgb]{1, 0.9725, 0.894} \sysname: ViT-S  & Int8                  & 41.0       & 40.5      & 40.8                    & 25.8       & 28.0      & 27.0                    \\
  \noalign{\vskip 2pt} 
  \hline
  \end{tabular}
  }
  \end{table}

We assessed the generalization performance of \sysname by training the models on ScanNet and evaluating their open-vocabulary classification accuracy on the unseen SUNRGBD and  NYU2 datasets: 
i. \textbf{SUNRGBD} dataset \cite{song2015sunrgbd} contains 5,285  training and 5,050 test RGBD images with 19 scene categories. The images are captured by multiple RGBD sensors, including Kinect, Asus Xtion, and Intel RealSense; and ii. \textbf{NYU2} dataset \cite{silberman2012nyuv} contains 795 training and 654 test RGBD images with 10 scene categories, which are captured by a Microsoft Kinect sensor.

As shown in \cref{tab:cross-dataset}, a trade-off is revealed: while \sysname-enhanced encoders significantly improved depth image accuracy (up to 25.4\% with DAT-T), there was a slight decrease in RGB image accuracy compared to the pre-trained CLIP model's ViT-G encoder. This was expected due to the substantial reduction in model size 
post-quantization 
(\eg, Swin-T's 56MB vs. CLIP-G's 5,213MB). 
Besides, CLIP's training involved 400 million image-text pairs, while \sysname used only 4,725 RGB and depth image pairs from ScanNet, suggesting that the limited adaptation dataset size may affect the model's generalization ability.

%% file: sections/conclusion.tex
\section{Conclusion}
\sysname showcases a significant advancement in leveraging a pre-trained visual-linguistic model for open-vocabulary classification across diverse image modalities, including both RGB and non-RGB domains. Despite its innovative approach, EdgeVL encounters a challenge in preserving generalization performance for RGB images when adapted for cross-modal use. Future work will concentrate on refining adaptation techniques to overcome this limitation, aiming to enhance the framework's versatility and effectiveness in a broader range of applications. 

%% file: sections/quantization.tex
In the supplemental material included, we offer further information regarding the quantization procedure and the approach to quantization-aware training implemented for the student model. Additionally, we outline the methodology utilized for creating label supersets with the assistance of ChatGPT-4 \cite{achiam2023gpt}. We also include experimental results concerning the overall efficacy of the student model, alongside an ablation study. To enhance comprehension of the model's effectiveness across various visual modalities, we conclude with visual representations of the open vocabulary classification outcomes on multiple datasets.

\section{Model Quantisation} 
We consider integer uniform quantization \cite{xiao2023smoothquant} which maps high-precision model weights and activations to low-precision representations. The quantization process is presented as follows:
\begin{equation}
    \begin{aligned}
        s &= \frac{2^{b-1}-1}{\alpha}\\
        \texttt{Quantization}: x_q &= \texttt{clip}(\lceil x \cdot s \rfloor, -2^{b-1}+1, 2^{b-1}-1)  \\
        \texttt{De-quantization}: \widetilde{x} &= \frac{x_q}{s}
    \end{aligned}
\end{equation}
where $\alpha$ is the maximum representable value, $s$ is the scale factor, $b$ is the bit-width, $x$ is the original weight or activation, $x_q$ is the quantized weight or activation, $\widetilde{x}$ is the weight or activation restored from quantized ones, and $\texttt{clip}$ is the clipping function. 
\begin{equation}
    \begin{aligned}
        \texttt{clip}(x, a, b) = \begin{cases}
            a & \text{if } x < a \\
            b & \text{if } x > b \\
            x & \text{otherwise}
        \end{cases}
    \end{aligned}
\end{equation}

\begin{figure}[tp]
    \centering
    \includegraphics[width=\linewidth]{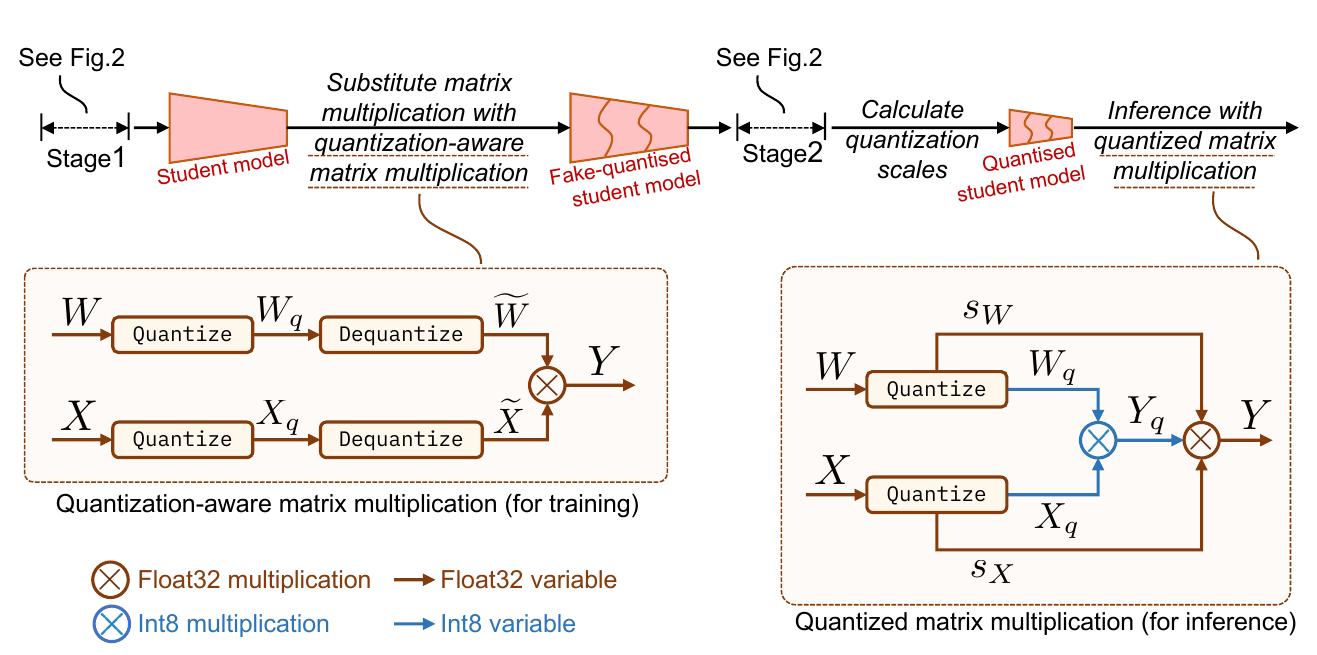}
    \caption{Quantization-aware matrix multiplication is used to take-quantize the student model}
    \label{fig:visual_examples}
\end{figure}

Per-tensor quantization applies the same scale factor across the entire tensor, whereas per-channel quantization assigns a distinct scale factor to each channel. The scale factor is determined using a calibration set in a static quantization approach or computed on the fly during inference for dynamic quantization. Quantizing the weights and activations of a Float32 model to Int8 reduces both the model size and inference latency. We follow the common practice \cite{wu2020integer}\cite{xiao2023smoothquant} and apply per-channel quantization to the weights and per-tensor quantization to the activations. When evaluating static quantization performance, we follow \cite{xiao2023smoothquant} and calculate the scale factors using a subset of the training set, \eg, 64 pairs of RGB and non-RGB images.

Figure \ref{fig:visual_examples} depicts the computation flow of quantization-aware matrix multiplication during training and quantized matrix multiplication during inference. After training the student model in stage-1, the fake-quantized student model is obtained by including the \texttt{Quantize} and \texttt{Dequantize} operations in the model. After stage-2, quantization scales are computed using a subset of the training data. For deployment on edge devices, the fake-quantized student model is transformed into the quantized student model using TensorRT, replacing quantization-aware matrix multiplication with quantized matrix multiplication.

%% file: sections/superset.tex
\section{Label Supersets Generation} 

Scene classification typically involves categorizing scenes within the domain of indoor, outdoor, or remote sensing. Acknowledging this, we create a comprehensive set of labels tailored to various scene classification challenges. For instance, we might develop a set of indoor scene labels for the ScanNet dataset and a distinct set of labels for satellite imagery in the EuroSAT dataset. In particular, we utilize ChatGPT-4 \cite{achiam2023gpt} to expertly craft these extensive label supersets.

For the ScanNet dataset, we use the following prompt:
\pagebreak
\begin{mdframed}[style=custombox]

\noindent \textbf{User}: Please generate a list of indoor scene categories without descriptions.

\noindent\hdashrule[0.5ex]{\linewidth}{0.7pt}{4pt 2pt}
\noindent ChatGPT-4: Living Room, Kitchen, Bedroom, Bathroom, Dining Room, Home Office, Basement, Attic, Office and Cubicles, Conference Room, Classroom, Laboratory, Library, Gym/Fitness Center, Shopping Mall, Retail Store, Hotel Lobby, Hotel Room, Restaurant, Cafe, Bar or Nightclub, Cinema/Theater, Airport Terminal, Subway Station, Train Station, Bus Interior, Hospital Ward, Operating Room, Dental Clinic, Veterinary Clinic, Museum, Art Gallery, Church, Temple, Mosque, Classroom, Lecture Hall, Laboratory, School Hallway, Dormitory, Indoor Pool, Bowling Alley, Arcade, Indoor Playground, Courthouse, City Hall, Police Station, Fire Station, Factory Floor, Warehouse, Server Room.

\end{mdframed}
For the EuroSAT dataset, we use the following prompt:
\begin{mdframed}[style=custombox]
    \noindent \textbf{User}: Please generate a list of satellite scene categories without descriptions.

    \noindent\hdashrule[0.5ex]{\linewidth}{0.7pt}{4pt 2pt}
    \noindent ChatGPT-4: Urban or Built-up Areas, Agriculture, Forest, Grassland, Water Bodies, Wetlands, Desert, Snow and Ice, Bare Land or Barren, Mountainous Terrain, Coastal Zones, Riparian Zones, Glacial and Periglacial Landscapes, Volcanic Features, Urban Green Spaces.
\end{mdframed}

\input{sections/longtable1.tex}

%% file: sections/longtable1.tex
\begin{table}[htp]
    \centering
    \renewcommand\arraystretch{1.2}
    \setlength\tabcolsep{4pt}  
    \caption{Overall accuracy comparison. \nrgb\ and \rgb\ denote the top1 accuracy of non-RGB and RGB images, respectively. And \avg\ denotes the average of the two, all in percentage. The same applies to the following tables. Best viewed in color.}
    \label{tab:overall_accuracy}
    \vspace{-5pt}
    \scalebox{0.9}{
      \begin{tabular}{ll|l|ccc|ccc} 
        \hline
        \noalign{\vskip 2pt} 
        \multirow{2}{*}{}       & \multirow{2}{*}{Methods~}                                              & \multicolumn{1}{c|}{\multirow{2}{*}{Bits}}                         & \multicolumn{3}{c|}{ScanNet (\%) $\uparrow$ }                                                                                                                                                                                                            & \multicolumn{3}{c}{EuroSAT (\%) $\uparrow$}                                                                                                                                                                                                              \\
        &                                                                        & \multicolumn{1}{c|}{}                                              & \nrgb                                                                       & \rgb                                                                        & \avg                                                                        & \nrgb                                                                       & \rgb                                                                        & \avg                                                                         \\ 
  
                                \noalign{\vskip 2pt} 
                                \hline
        \noalign{\vskip 2pt} 
                                & Pretrained CLIP-B \cite{radford2021clip}  & F32                                  & 4.5                                           & 36.2                                          & 20.4                                          & 16.8                                          & 40.4                                          & 28.6                                           \\
                                & Pretrained CLIP-G \cite{radford2021clip}  & F32                                  & 6.2                                           & 47.3                                          & 26.8                                          & 16.9                                          & 54.0                                          & 35.5                                           \\ 
                                \noalign{\vskip 2pt} 
                                \hline
        \noalign{\vskip 2pt} 
        \multirow{8}{*}{DAT-T}  & Frank \cite{hafner2022crossmodal}         & F32                                  & 8.3                                           & 21.7                                          & 15.0                                          & 32.4                                          & 31.4                                          & 31.9                                           \\
                                & Gupta \cite{hoffman2016crossmodal}        & F32                                  & 16.8                                          & 22.4                                          & 19.6                                          & 25.4                                          & 27.2                                          & 26.3                                           \\
                                & CMKD \cite{hong2022crossmodality} (non-RGB) & F32                                  & 37.8                                          & 11.5                                          & 24.6                                          & 36.7                                          & 18.9                                          & 27.8                                           \\
                                & CMKD \cite{hong2022crossmodality} (RGB)   & F32                                  & 4.0                                           & 42.5                                          & 23.2                                          & 15.2                                          & 40.9                                          & 28.1                                           \\
                                & Fida \cite{thoker2019crossmodal}          & F32                                  & 38.9                                          & 5.8                                           & 22.3                                          & 53.7                                          & 14.1                                          & 33.9                                           \\
                                & CQD \cite{su2017adapting}                 & F32                                  & 40.1                                          & 6.7                                           & 23.4                                          & 37.2                                          & 17.5                                          & 27.3                                           \\
                                & SKD \cite{yang2022mixskd}                 & F32                                  & 21.2                                          & 40.7                                          & 31.0                                          & 24.8                                          & 41.5                                          & 33.2                                           \\
                                & {\cellcolor[rgb]{1,0.973,0.894}}\sysname  & {\cellcolor[rgb]{1,0.973,0.894}}Int8 & {\cellcolor[rgb]{1,0.973,0.894}}\textbf{47.9} & {\cellcolor[rgb]{1,0.973,0.894}}\textbf{52.0} & {\cellcolor[rgb]{1,0.973,0.894}}\textbf{49.9} & {\cellcolor[rgb]{1,0.973,0.894}}\textbf{61.0} & {\cellcolor[rgb]{1,0.973,0.894}}\textbf{65.7} & {\cellcolor[rgb]{1,0.973,0.894}}\textbf{63.3}  \\ 
                                \noalign{\vskip 2pt} 
                                \hline
        \noalign{\vskip 2pt} 
        \multirow{8}{*}{Swin-T} & Frank \cite{hafner2022crossmodal}         & F32                                  & 8.0                                           & 14.8                                          & 11.4                                          & 41.9                                          & 39.0                                          & 40.5                                           \\
                                & Gupta \cite{hoffman2016crossmodal}        & F32                                  & 22.0                                          & 17.5                                          & 19.8                                          & 45.8                                          & 39.5                                          & 42.6                                           \\
                                & CMKD \cite{hong2022crossmodality} (non-RGB) & F32                                  & 38.9                                          & 4.0                                           & 21.4                                          & 59.5                                          & 22.0                                          & 40.8                                           \\
                                & CMKD \cite{hong2022crossmodality} (RGB)   & F32                                  & 3.2                                           & 42.4                                          & 22.8                                          & 20.1                                          & 62.4                                          & 41.2                                           \\
                                & Fida \cite{thoker2019crossmodal}          & F32                                  & 41.2                                          & 1.3                                           & 21.3                                          & 54.5                                          & 10.5                                          & 32.5                                           \\
                                & CQD \cite{su2017adapting}                 & F32                                  & 41.4                                          & 4.9                                           & 23.1                                          & \textbf{62.5}                                          & 18.3                                          & 40.4                                           \\
                                & SKD \cite{yang2022mixskd}                 & F32                                  & 31.2                                          & 37.8                                          & 34.5                                          & 24.1                                          & 28.8                                          & 26.4                                           \\
                                & {\cellcolor[rgb]{1,0.973,0.894}}\sysname  & {\cellcolor[rgb]{1,0.973,0.894}}Int8 & {\cellcolor[rgb]{1,0.973,0.894}}\textbf{46.0} & {\cellcolor[rgb]{1,0.973,0.894}}\textbf{48.7} & {\cellcolor[rgb]{1,0.973,0.894}}\textbf{47.4} & {\cellcolor[rgb]{1,0.973,0.894}}61.3          & {\cellcolor[rgb]{1,0.973,0.894}}\textbf{67.1} & {\cellcolor[rgb]{1,0.973,0.894}}\textbf{64.2}  \\ 
                                \noalign{\vskip 2pt} 
                                \hline
        \noalign{\vskip 2pt} 
  
        \multirow{8}{*}{ViT-S}  & Frank \cite{hafner2022crossmodal}         & F32                                  & 11.3                                          & 14.0                                          & 12.7                                          & 49.2                                          & 37.9                                          & 43.5                                           \\
                                & Gupta \cite{hoffman2016crossmodal}        & F32                                  & 19.6                                          & 14.9                                          & 17.2                                          & 54.2                                          & 42.4                                          & 48.3                                           \\
                                & CMKD \cite{hong2022crossmodality} (non-RGB) & F32                                  & 38.0                                          & 4.7                                           & 21.3                                          & 61.2                                          & 34.4                                          & 47.8                                           \\
                                & CMKD \cite{hong2022crossmodality} (RGB)   & F32                                  & 3.0                                           & 42.5                                          & 22.8                                          & 14.3                                          & 60.4                                          & 37.3                                           \\
                                & Fida \cite{thoker2019crossmodal}          & F32                                  & 40.0                                          & 4.5                                           & 22.2                                          & 56.7                                          & 20.3                                          & 38.5                                           \\
                                & CQD \cite{su2017adapting}                 & F32                                  & 38.0                                          & 4.0                                           & 21.0                                          & 62.4                                          & 36.4                                          & 49.4                                           \\
                                & SKD \cite{yang2022mixskd}                 & F32                                  & 28.7                                          & 37.6                                          & 33.1                                          & 22.9                                          & 50.3                                          & 36.6                                           \\
                                & {\cellcolor[rgb]{1,0.973,0.894}}\sysname  & {\cellcolor[rgb]{1,0.973,0.894}}Int8 & {\cellcolor[rgb]{1,0.973,0.894}}\textbf{42.0} & {\cellcolor[rgb]{1,0.973,0.894}}\textbf{47.5} & {\cellcolor[rgb]{1,0.973,0.894}}\textbf{44.7} & {\cellcolor[rgb]{1,0.973,0.894}}\textbf{62.9} & {\cellcolor[rgb]{1,0.973,0.894}}\textbf{66.8} & {\cellcolor[rgb]{1,0.973,0.894}}\textbf{64.8}  \\
                                \noalign{\vskip 2pt} 
                                \hline
        \end{tabular}
    }
  \end{table}

%% file: sections/results.tex
\section{Results}

\subsection{Overall Accuracy}

To provide a comprehensive performance evaluation of \sysname\ compared to existing methods, we present a detailed accuracy analysis in \cref{tab:overall_accuracy}. Except for a small performance difference to CQD \cite{su2017adapting} using Swin-T on EuroSAT (61.3\% vs. 62.5\%), \sysname\ demonstrates superior performance across all backbones and on two datasets, surpassing all other methods by a significant margin.

\subsection{Quantization Aware Contrastive Learning}
\input{sections/longtable2.tex}

We compare the performance of \sysname\ with other quantization methods when dynamic quantization is applied during inference. Comparison between Tab. 3 in the main script and \cref{tab:open-vocabulary-classification-quantized} reveals that dynamic quantization outperforms static quantization. This finding is also supported by \cite{xiao2023smoothquant}. Just like the results showcased in Tab. 3 in the main script, \sysname\ emerges as the most effective method in maintaining model performance post-quantization.

\subsection{Effect of Finetuning CLIP}
To further demonstrate the effectiveness of our fine-tuning method, we use the CLIP-B model as the student model and the CLIP-G model as the teacher model. The CLIP-B model was trained following the same approach as in stage-1 of Fig. 1, with training parameters identical to those in Sec. 4.1. As shown in \cref{tab:finetuned_clip}, the fine-tuned CLIP-B model exhibits improved performance compared to the pre-trained models on non-RGB modalities, but its performance still lags much behind \sysname. We attribute this to the fact that the CLIP-B model is designed for large-scale image-text retrieval tasks, meaning it excels with larger datasets. In contrast, \sysname\ is specifically optimized for cross-modal classification tasks. Additionally, the contrastive learning process during stage 2 helps \sysname\ to learn better feature representations.

\begin{table}[htp]
  \centering
  \renewcommand\arraystretch{1.2}
  \setlength\tabcolsep{4pt}  
  \caption{Effect of finetuning CLIP-B.}
  \label{tab:finetuned_clip}
  \vspace{-8pt}
  \scalebox{0.8}{
  \begin{tabular}{l|l|ccc|ccc} 
  \hline
  \noalign{\vskip 2pt} 
  \multirow{2}{*}{Methods}                            & \multirow{2}{*}{Bits} & \multicolumn{3}{c|}{ScanNet (\%) $\uparrow$ }                        & \multicolumn{3}{c}{EuroSAT (\%) $\uparrow$}                          \\ 
                                                      &                       & \nrgb      & \rgb       & \avg            & \nrgb      & \rgb       & \avg             \\ 
                                                      \noalign{\vskip 2pt} 
                                                      \hline
  \noalign{\vskip 2pt} 
  Finetuned CLIP-B                   & F32                   & 27.1          & 27.9          & 27.5          & 46.9          & 47.7          & 47.3           \\
\rowcolor[rgb]{1, 0.9725, 0.894} \sysname (DAT-T)  & Int8                  & \textbf{47.9} & \textbf{52.0} & \textbf{49.9} & \textbf{61.0}          & \textbf{65.7}          & \textbf{63.3}           \\

  \noalign{\vskip 2pt} 
  \hline
  \end{tabular}
  }
  \vspace{-10pt}

\end{table}

\subsection{Qualitative Analysis}

\begin{figure}[htp]
  \centering
  \includegraphics[width=0.9\linewidth]{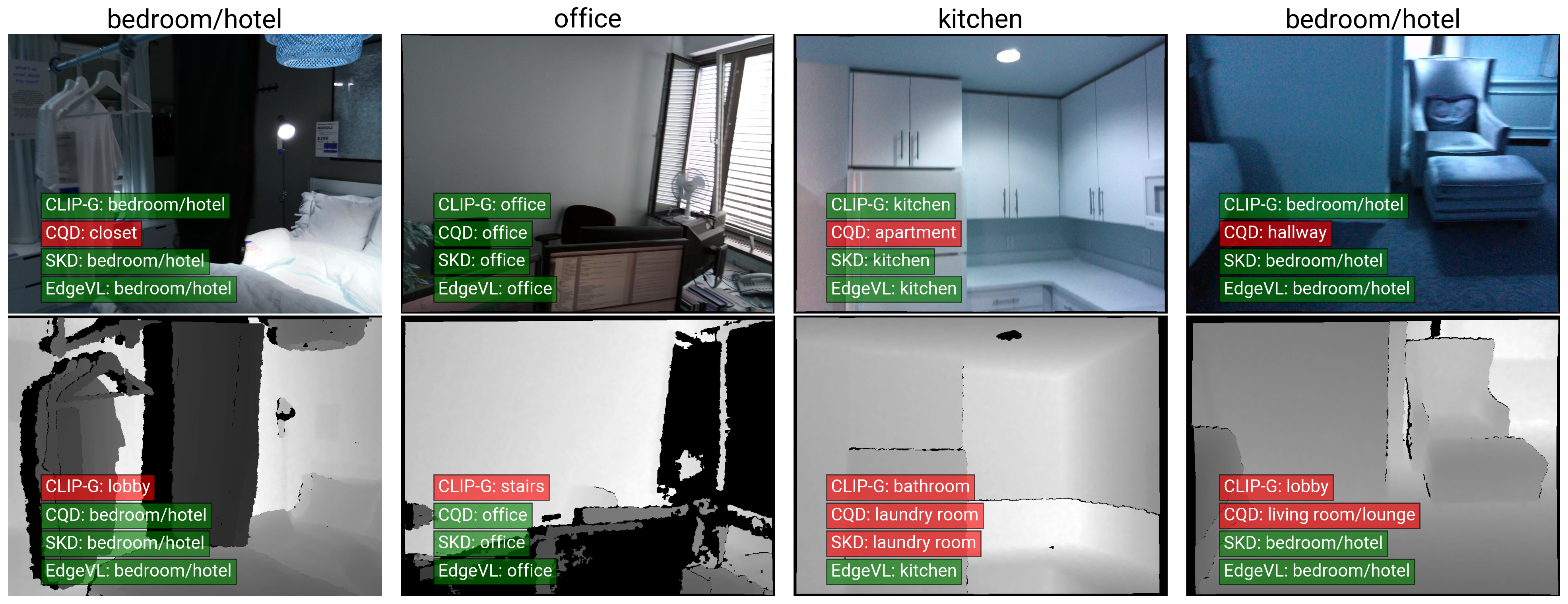}
  \caption{Visualization of the predictions of different models on ScanNet. Red and green colors indicate incorrect and correct predictions, respectively.}
  \label{fig:visual_examples_scannet}
\end{figure}

\begin{figure}[htp]
\centering
\includegraphics[width=0.9\linewidth]{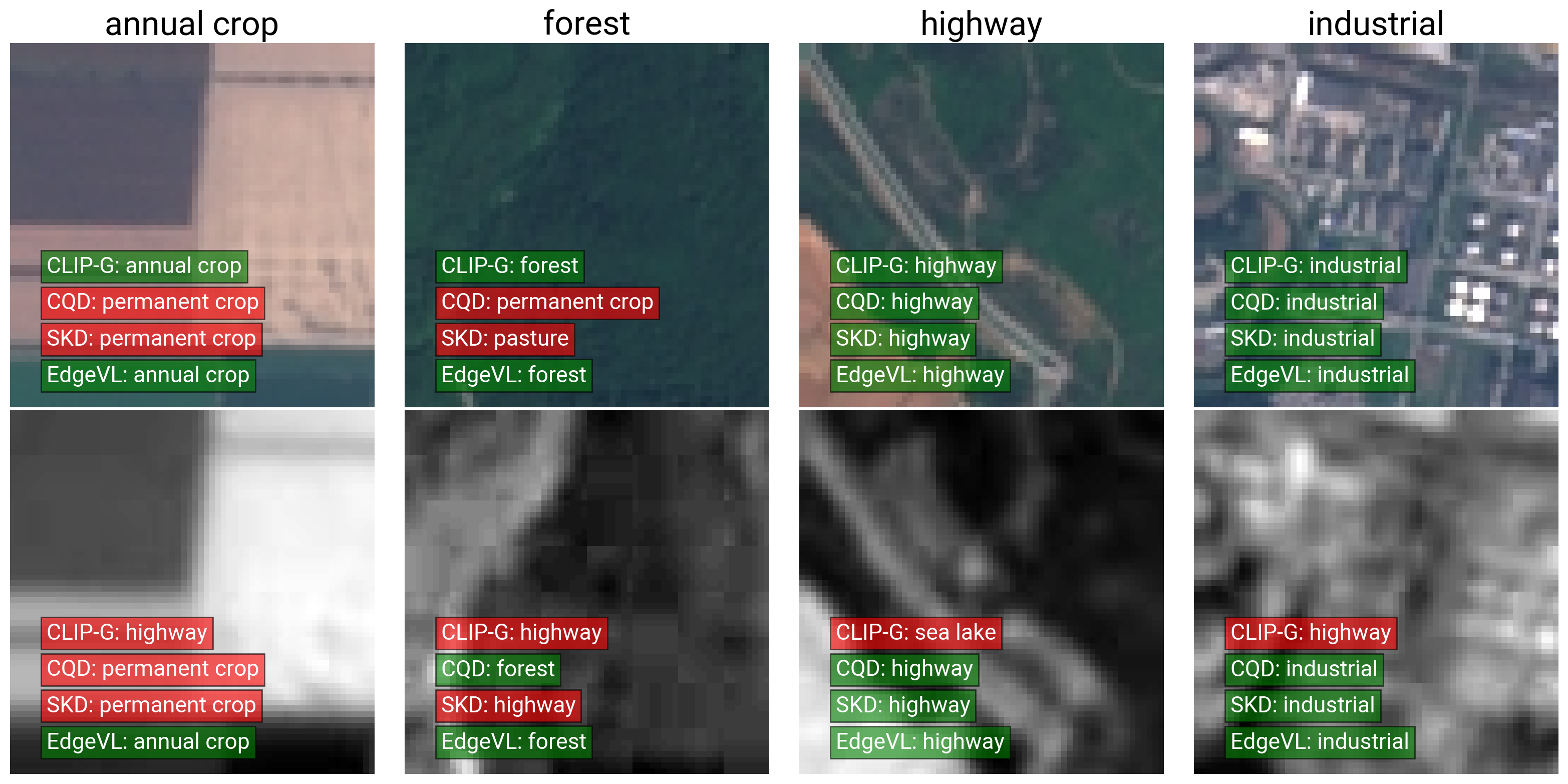}
\caption{Visualization of the predictions of different models on EuroSAT. Red and green colors indicate incorrect and correct predictions, respectively.}
\label{fig:visual_examples_eurosat}
\end{figure}

\begin{figure}[htp]
\centering
\includegraphics[width=0.9\linewidth]{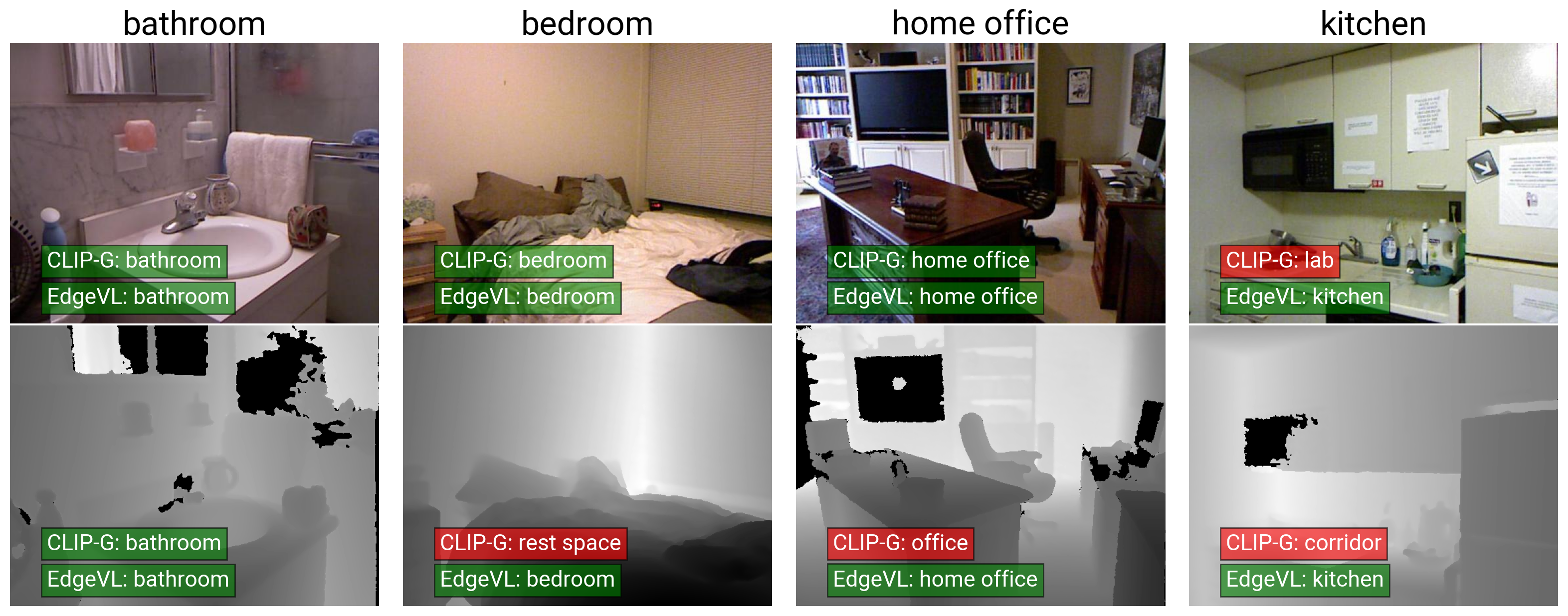}
\caption{Visualization of the predictions of different models on SUNRGBD. Red and green colors indicate incorrect and correct predictions, respectively.}
\label{fig:visual_examples_sunrgbd_good}
\end{figure}

\begin{figure}[htp]
  \centering
  \includegraphics[width=0.9\linewidth]{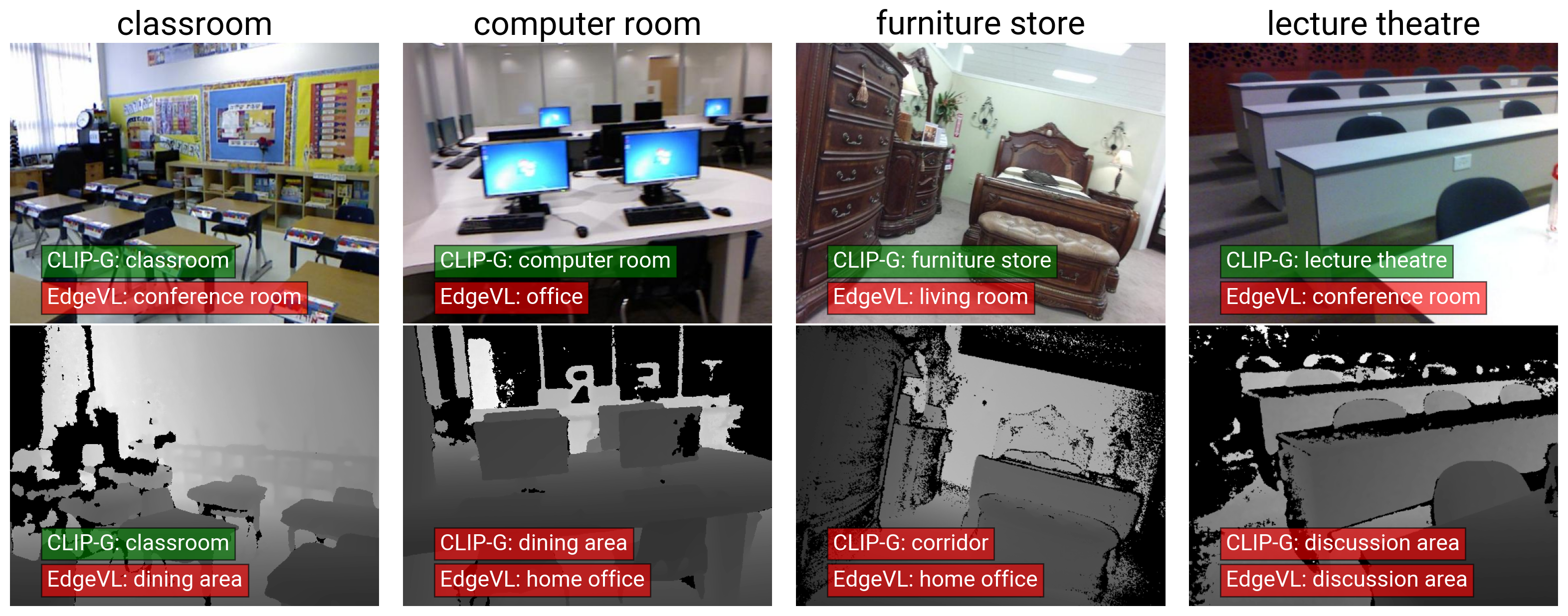}
  \caption{Visualization of the predictions of different models on SUNRGBD. Red and green colors indicate incorrect and correct predictions, respectively.}
  \label{fig:visual_examples_sunrgbd_bad}
  \end{figure}

\Cref{fig:visual_examples_scannet} and \Cref{fig:visual_examples_eurosat} illustrate visual examples of the predictions made by CLIP-G \cite{radford2021clip}, CQD \cite{su2017adapting}, SKD \cite{yang2022mixskd}, and \sysname\ (ViT-S) on the ScanNet and EuroSAT datasets, respectively. The visualizations underscore the superior accuracy of \sysname\ over competing methods, particularly with non-RGB images.

\Cref{fig:visual_examples_sunrgbd_good} and \Cref{fig:visual_examples_sunrgbd_bad} depict the cross-dataset predictions on SUNRGBD by \sysname\ (DAT-T) that was trained on ScanNet. In \Cref{fig:visual_examples_sunrgbd_good}, \sysname\ demonstrates good performance in distinctive scenes like bathrooms and kitchens. However, in settings such as furniture stores and lecture theatres shown in \Cref{fig:visual_examples_sunrgbd_bad}, \sysname\ struggles even with the RGB modality, while CLIP-G achieves higher accuracy in these instances. This is due to the infrequent presence of these scenarios (e.g., furniture stores and lecture theatres) in the ScanNet dataset, making them unfamiliar to the model. We also discussed the potential reason for data amount in our main paper. Our future research will focus on improving the model's generalization ability with limited adapataion data.

%% file: sections/longtable2.tex
\begin{table}[htp]
  \centering
  \renewcommand\arraystretch{1.2}
  \setlength\tabcolsep{4pt}  
  \caption{QAT meets Contrastive Learning. + denotes in combination with stage-1. Inferencing with Dynamic quantization.}
  \vspace{-5pt}
  \label{tab:open-vocabulary-classification-quantized}
  \scalebox{0.9}{
    \begin{tabular}{l|l|ccl|ccl|ccl} 
      \hline
      \noalign{\vskip 2pt} 
      \multirow{2}{*}{Methods}                        & \multirow{2}{*}{Bits} & \multicolumn{3}{c|}{DAT-T (\%)}                                & \multicolumn{3}{c|}{Swin-T (\%)}                               & \multicolumn{3}{c}{ViT-S (\%)}                                 \\
                                                      &                       & \nrgb         & \rgb          & \multicolumn{1}{c|}{\avg} & \nrgb         & \rgb          & \multicolumn{1}{c|}{\avg} & \nrgb         & \rgb          & \multicolumn{1}{c}{\avg}  \\ 
                                                      \noalign{\vskip 2pt} 
                                                      \hline
      \noalign{\vskip 2pt} 
      Stage-1                                         & F32                   & 38.6          & 40.6          & 39.6                      & 39.9          & 41.2          & 40.5                      & 37.8          & 40.7          & 39.3                      \\ 
      \noalign{\vskip 2pt} 
      \hline
      \noalign{\vskip 2pt} 
      +PTQ~\cite{jacob2018quantization}               & Int8                  & 35.1          & 39.1          & 37.1                      & 32.0          & 35.0          & 33.5                      & 26.6          & 28.5          & 27.6                      \\
      +QAT \cite{jacob2018quantization}               & Int8                  & 39.6          & 41.3          & 40.5                      & 39.6          & 39.5          & 39.5                      & 38.4          & 41.9          & 40.1                      \\
      +QViT~\cite{li2022qvit}                         & Int8                  & 38.8          & 41.1          & 39.9                      & 37.8          & 39.9          & 38.8                      & 31.9          & 35.2          & 33.5                      \\
      \rowcolor[rgb]{1,0.973,0.894} \textbf{+Stage-2} & Int8                  & \textbf{49.0} & \textbf{51.5} & \textbf{50.2}             & \textbf{47.6} & \textbf{49.7} & \textbf{48.7}             & \textbf{44.9} & \textbf{49.1} & \textbf{47.0}             \\
      \noalign{\vskip 2pt} 
      \hline
      \end{tabular}
  }
\end{table}